\documentclass[twoside,11pt]{article}

\usepackage{jmlr2e, subfigure, amsmath, algorithm, algpseudocode}

\PassOptionsToPackage{hyphens}{url}\usepackage[hidelinks]{hyperref}
\makeatletter
\renewcommand\hyper@natlinkbreak[2]{#1}
\makeatother

\algnewcommand{\LComment}[1]{\hspace{0.3cm}\(\triangleright\) #1}

\jmlrheading{}{}{}{}{}{}

\ShortHeadings{3-D Hand Pose Estimation from Kinect's Point Cloud}{Coscia, Palmieri, Castaldo and Cavallo}
\firstpageno{1}

\begin{document}

\title{3-D Hand Pose Estimation from Kinect's Point Cloud Using Appearance Matching}

\author{\name Pasquale Coscia \email pasquale.coscia@unina2.it \\
\name Francesco A.N. Palmieri \email francesco.palmieri@unina2.it \\
\name Francesco Castaldo \email francesco.castaldo@unina2.it \\
\name Alberto Cavallo \email alberto.cavallo@unina2.it \\
       \addr Seconda Universit\'a di Napoli (SUN),\\
        Dipartimento di Ingegneria Industriale e dell'Informazione,\\
         via Roma 29, 81030 Aversa (CE) - Italy
}

\maketitle

\begin{abstract}
We present a novel appearance-based approach for pose estimation of a human hand using the point clouds provided by the low-cost Microsoft Kinect sensor. Both the free-hand case, in which the hand is isolated from the surrounding environment, and the hand-object case, in which the different types of interactions are classified, have been considered. The hand-object case is clearly the most challenging task  having to deal with multiple tracks. The approach proposed here belongs to the class of partial pose estimation where the estimated pose in a frame is used for the initialization of the next one. The pose estimation is obtained by applying a modified version of the Iterative Closest Point (ICP) algorithm to synthetic models to obtain the rigid transformation that aligns each model with respect to the input data. The proposed framework uses a ``pure'' point cloud as provided by the Kinect sensor without any other information such as RGB values or normal vector components. For this reason, the proposed method can also be applied to data obtained from other types of depth sensor, or RGB-D camera. 
\end{abstract}

\begin{keywords}
Hand Pose, Kinect, Point Cloud, RANSAC, DBSCAN, ICP
\end{keywords}

\section{Introduction}
\label{sec:introduction}
Hand pose estimation, by means of one or more cameras, is a desired feature of many information systems in several applications both in navigation and manipulation. For example in robotic applications, where robots are  equipped with dexterous hands, it is important to provide feedback to the robot itself and/or to qualify how well it may be performing human-like grasps. In manipulators,  efficient processing for analyzing hand motions may be an important part of a system specially when a device has to control complex machinery with many degrees of freedom (DoFs). 

Currently, the best results for capturing hand motion are obtained by means of visual markers, electro-mechanical/magnetic sensing devices, or other specifically designed hardware. They provide a complete set of measurements for all hand's functionalities. Unfortunately, they have several disadvantages in term of casual use as they are invasive, expensive, restrict the naturalness  of motion and require complex calibration and setup  in order to obtain accurate measurements. Therefore, there is a great interest in developing markerless based solutions because of their potential to provide more natural interaction. Despite the significant progress in this field, the problem remains still open and reveals various theoretical and practical challenges due to a large number of issues. Fundamentally, the main problems encountered in the design of hand pose estimation systems include: high-dimensionality, self-occlusions,  speed processing and hand speed. \\As stated in Erol et al. \citep{Erol200752}, there are two main approaches to the hand pose estimation problem: \emph{partial} and \emph{full} DoF. \emph{Partial} means that we do not have all the kinematics parameters of the hand's skeleton, but only rough information about the hand motion, such as position of the palm or the fingers. Contrariwise, \emph{full} DoF pose estimation approaches attempt to estimate all the kinematic parameters of the hand skeleton.\\ The possibility to use depth information with a relatively low-cost sensors has given new perspectives to the  hand tracking. Among all, the Kinect sensor \citep{Kinect_Zhang} has become one of the most used depth sensor. The development of different versions of such device over the years, with increasingly high performance, proves its importance in the fields of gaming and research. It combines two technology, a RGB camera and an IR camera.  To obtain depth information it uses the structured light principle. The idea behind this principle is to project a known pattern onto the scene and infer the depth of the objects using the deformation of such pattern and a triangulation method.\\ The Kinect has pioneered the development of new types of depth sensors such as the 3Gear Systems interface \citep{sensor:3gear}, the Leap Motion controller \citep{sensor:leapmotion} or the most recent Creative Senz3D camera \citep{sensor:senz3d}. The 3Gear Systems interface is able to estimate hand poses using two Kinect cameras. The Leap Motion technology enables users to manipulate virtual objects with hand motions. The Creative Senz3D performs a close-range hand and finger tracking as well as facial analysis. However, more research effort is needed to understand the real potentialities of these devices.\\ The output of depth sensors is typically a point cloud that has become one of the most common framework for object representation in computer vision. A point cloud is a data structure containing a set of multi-dimensional points expressed within a given  coordinate system and is typically used to represent the external surface of an object. In a three-dimensional coordinate system, these points usually stand for the \emph{x}, \emph{y} and \emph{z} geometric coordinates of the sampled surface. Each point may also have other attributes such as the components of the normal vector, the accuracy, or the color in the four spectral bands.  In this work only the \emph{xyz}-coordinates of the point clouds have been used. \\ The main contribution of this paper is a method, based on a combination of computer vision algorithms, for estimating  human hand pose during different static gestures as well as predefined grasping tasks. Throughout the paper different methods and algorithms are carefully selected to achieve the best results in terms of reliability and robustness of the recognition task. \\The rest of the paper is organized as follows: Section \ref{sec:related_work} provides a brief overview of the state of the art regarding the hand pose estimation problem. Section \ref{sec:the_approach} presents in detail each step of the proposed framework and their relative outputs. Section \ref{sec:exp_results} shows experimental results for a number of real test sequences. The last section discusses the results of the proposed framework and suggestions for improvements.  

\section{Related work}
\label{sec:related_work}
The 3-D hand pose estimation problem has been addressed by many authors, including Stenger et al. \citep{990976}, Rehg and Kanade \citep{Rehg:1994}, de La Gorce \citep{5719617} for a single RGB camera and de Campos and Murray \citep{1640833} for multiple cameras. Athitsos et al. \citep{1211500} estimate 3-D hand pose from cluttered images. They formulate the hand pose estimation as an image indexing problem, using a large database of synthetic hand images. They also use an approximation of the image-to-model chamfer distance and a probabilistic line matching method to retrieve the closest matches for an input hand from the database. Rosales et al. \citep{937543} propose a system for recovering 3-D hand pose from monocular color sequences. Their system employs a non-linear supervised learning framework where the training set is obtained via a CyberGlove. \\Several approaches have also been proposed to estimate the hand pose from depth images. Mo et al. \citep{1640934} use a laser-based camera to produce low-resolution depth images. The method is based on recognizing hand poses as finger states. Malassiotis et al. \citep{Malassiotis20081027} extract PCA features from depth images of synthetics 3-D hand models for training. Liu et Fujimura \citep{1301587} recognize hand gestures by using a sequence of real-time depth image data acquired by an active sensing hardware. Their approach makes full use of 3-D trajectory as well as shape information for classifying gestures. Suryanarayan et al. \citep{5597253} use depth information to recognize scale and rotation invariant hand poses dynamically. They have designed a volumetric shape descriptor enfolding the hand to generate a 3-D cylindrical histogram and to achieve a real-time pose recognition.\\ A milestone is represented by the work of Oikonomidis et al. \citep{BMVC.25.101}. They propose a tracking-based method by making use of a RGB-D Kinect camera. They treat the pose estimation as an optimization problem, using a variant of Particle Swarm Optimization (PSO), selecting parameters that minimize the error between the 3-D hypothesized instances and visual observations of the hand. The method, however, requires an explicit initialization step for better results. They also assert that their algorithm can run at about 15 FPS if it is implemented on a GPU that corresponds to only half the rate at which the Kinect can operate. More recently, the approach proposed by Melax et al. \citep{Melax:2013} for 3-D markerless hand tracking from a depth camera uses an augmented rigid body simulation that allows to handle tracking as a linear complementary problem. To limit motion, the system generates constraints that are resolved with a projected Gauss-Seidel solver. \\ Different approaches have also been proposed to deal with the interaction with objects. In particular, Fu and Santello \citep{6092033} have created a framework for tracking hand and object during grasping tasks. Grasping tasks are modeled as a collision detection problem while the hand is tracked with a marker-based solution using an Extended Kalman Filter. Hamer et al. \citep{5540150} address the problem of manipulating objects using individual local tracker for each part of an articulated structure. A collection of surfaces patches forms the hand and a Markov random field enforces valid configurations connecting the segments.

\section{Our Approach}
\label{sec:the_approach}

\begin{figure}[]
  \centering
  \includegraphics[scale=0.80]{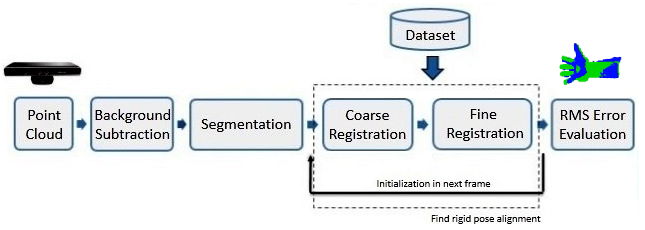}\\
  \caption{Overview of the proposed framework for 3-D hand pose estimation.}\label{system}
\end{figure}

Fig. \ref{system} shows the block diagram of the proposed framework. The first step is background subtraction followed by clustering and matching steps  in order to estimate the hand pose. Since the output of the matching step is fed back for the next frame, there is no real tracking. The starting point of the search, at each frame, is based on the result of the previous one. The pose is represented as a frame-dependent rotation matrix \textbf{R}[k] and a translation vector \textbf{t}[k] that update the position of the models' point clouds $\textbf{M}_i$ in the space: $\textbf{M}_i[k] = \textbf{R}[k] \cdot \textbf{M}_i[k-1] + \textbf{t}[k]$. 

Our case study consists of a scene in which the hand is located in front of the sensor and has a plane background. For grasping tasks, it has been considered only one object at time, situated on a table. The hand and the objects are placed at a fixed distance. Therefore there is no need to change the scale of the model. The point clouds have been stored at approximately 30 Hz and then processed. Each of them, before the processing, contains approximately 200k 3-D points. The point clouds have been captured using the Simulink Support for Kinect \citep{ssfk} and all the algorithms have been written in MATLAB.

\subsection{Background Subtraction}
\label{subsec:background_subtraction}
The first step in processing the data consists in removing the background, that in our case was roughly a plane (a wall). To perform the subtraction, we used the RANSAC algorithm. RANSAC stands for `'RANdom SAmple Consensus'' and it is an iterative method for a robust  parameters estimation  of a mathematical model from a set of data containing outliers. In our case the model is represented by a plane, while outliers are the hand and the objects. It is a nondeterministic algorithm because it produces the correct result only with a given probability, which raises with the increase of the number of iterations. Its main applications include 3-D models fitting, stereo matching and linear regression. Several variations of the algorithm have been proposed  using different cost functions or adaptive parameters for better performance \citep{ransac1, ransac2}. In this work, we used the function written by Kovesi \citep{KovesiMATLABCode}, that uses the RANSAC algorithm to robustly fit a plane to a set of 3D data points. The algorithm operates as follows:

\begin{itemize}
  \item {Randomly select a subset $\textbf{S}$ of the original data $\textbf{P}$.}
  \item {Estimate the model from the subset $\textbf{S}$.}
  \item {Test all data points $p_i\in \textbf{P}$ that fit the estimated model, i.e., they are within a distance threshold $\tau$ with respect to the model. Those points represent the \emph{consensus set}.}
  \item {Re-estimate the model using all points of the \emph{consensus set} if the number of inliers is greater than a predefined threshold, otherwise select a new subset $\textbf{S}$ of points and estimate a new model.}
  \item {After N trials, among all \emph{consensus set}, the largest set is selected.}
\end{itemize}

\begin{figure}[H]
\centering
\subfigure[]{\includegraphics[scale=0.20]{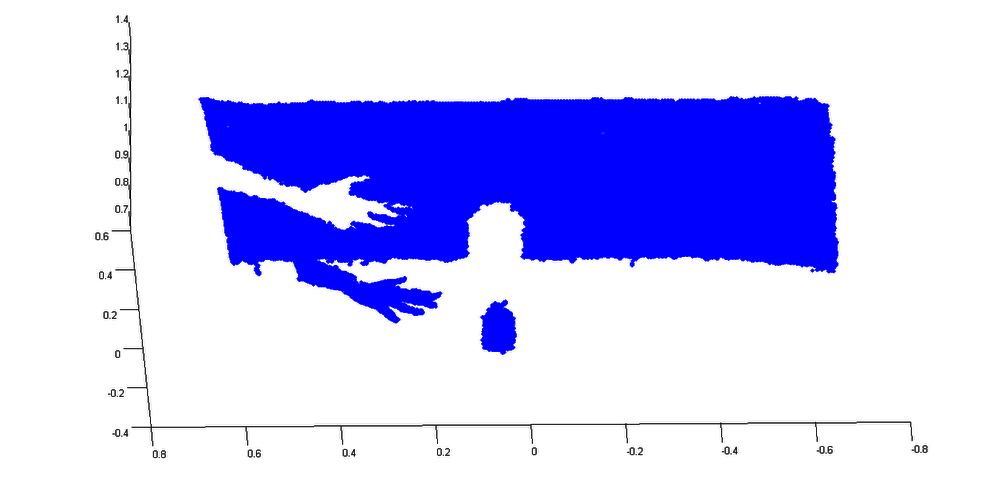}\label{ransac_result1}}
\subfigure[]{\includegraphics[scale=0.20]{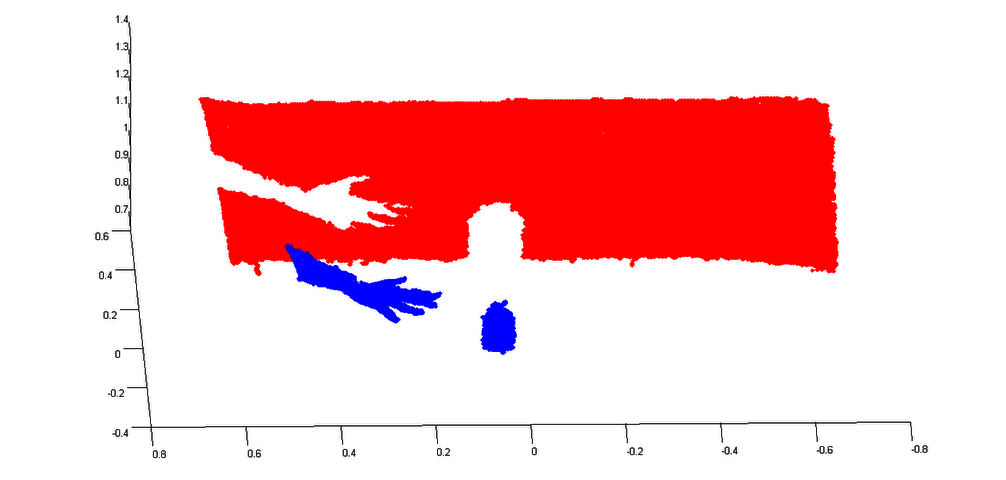}\label{ransac_result2}}
\subfigure[]{\includegraphics[scale=0.20]{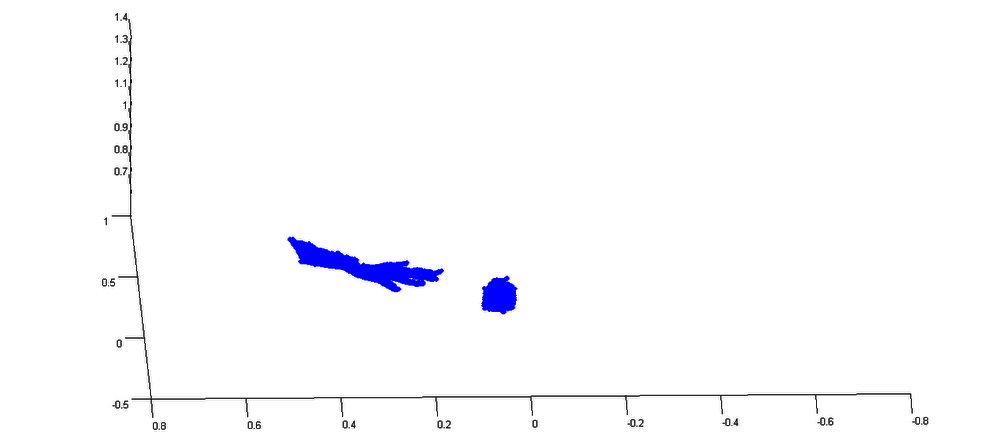}\label{ransac_result3}}
\caption[Results of applying RANSAC plane fitting algorithm.]{Results of applying RANSAC plane fitting algorithm: a) Original Point Cloud; b) Points (in red) that belong  to the plane model determined by RANSAC;  c) Point Cloud after background subtraction.}
\label{ransac_result123}
\end{figure}

\begin{figure}
\centering
\subfigure[]{\includegraphics[width=75mm,height=38mm]{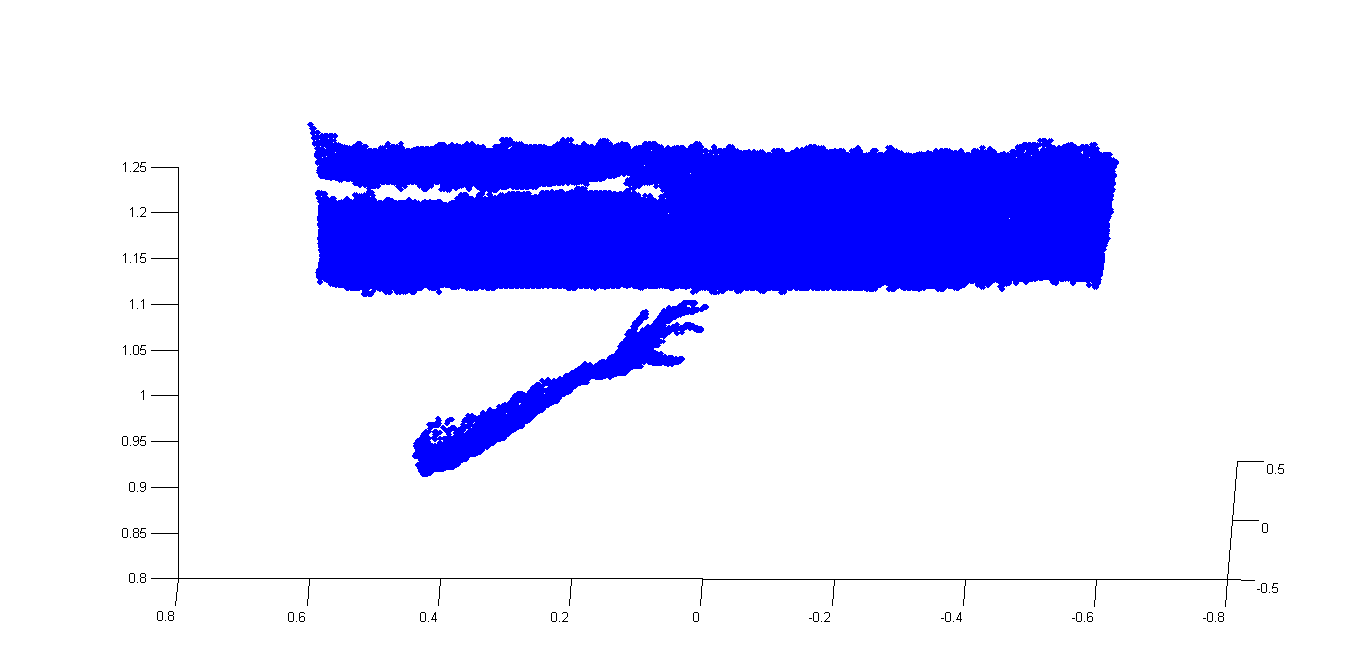}\label{ransac_th1}}\hspace{-1.3em}
\subfigure[]{\includegraphics[width=75mm,height=32mm]{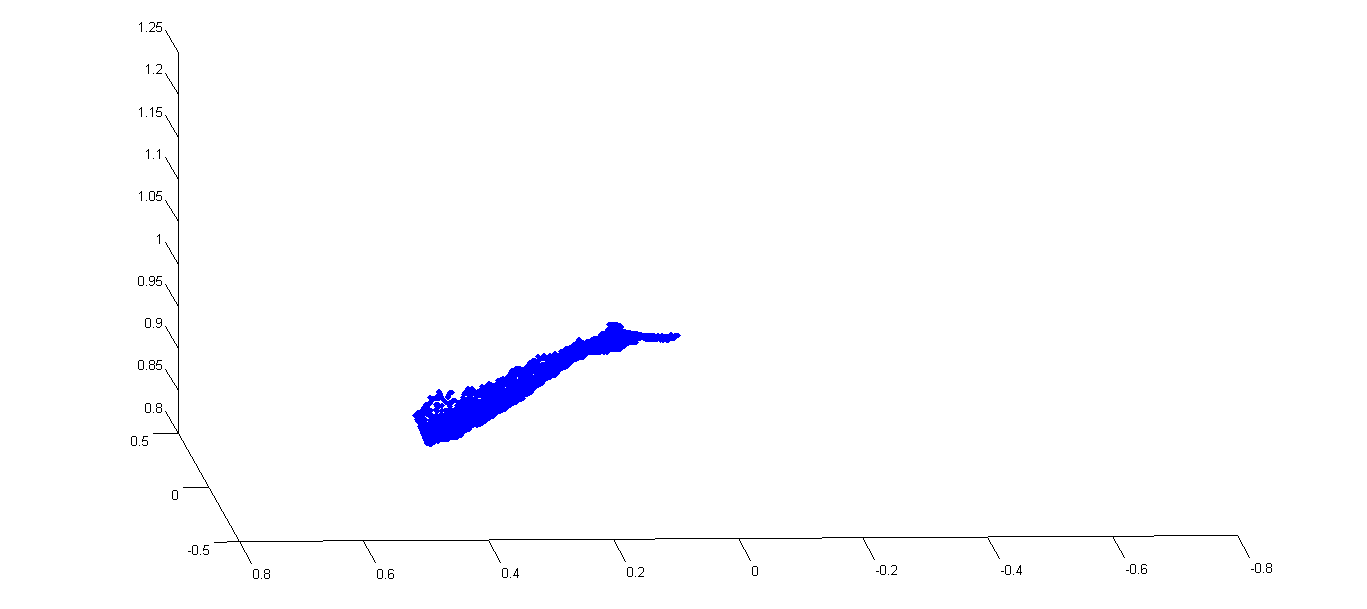}\label{ransac_th2}}
\subfigure[]{\includegraphics[width=75mm,height=32mm]{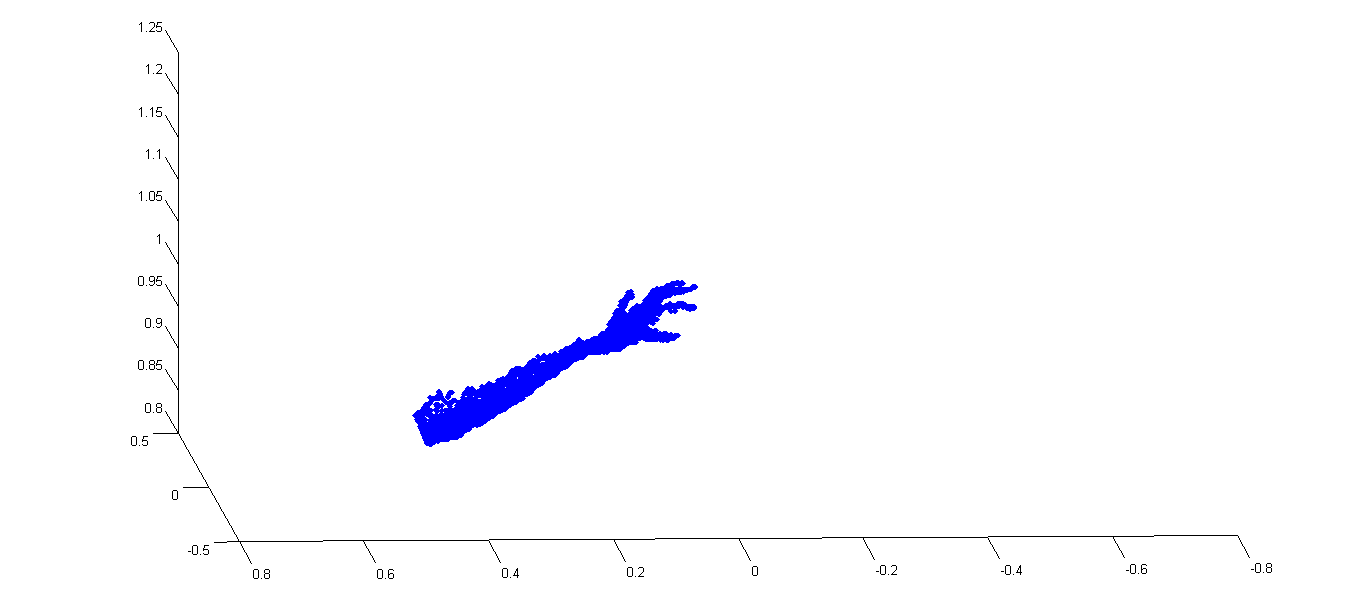}\label{ransac_th3}}\hspace{-1.3em}
\subfigure[]{\includegraphics[width=75mm,height=32mm]{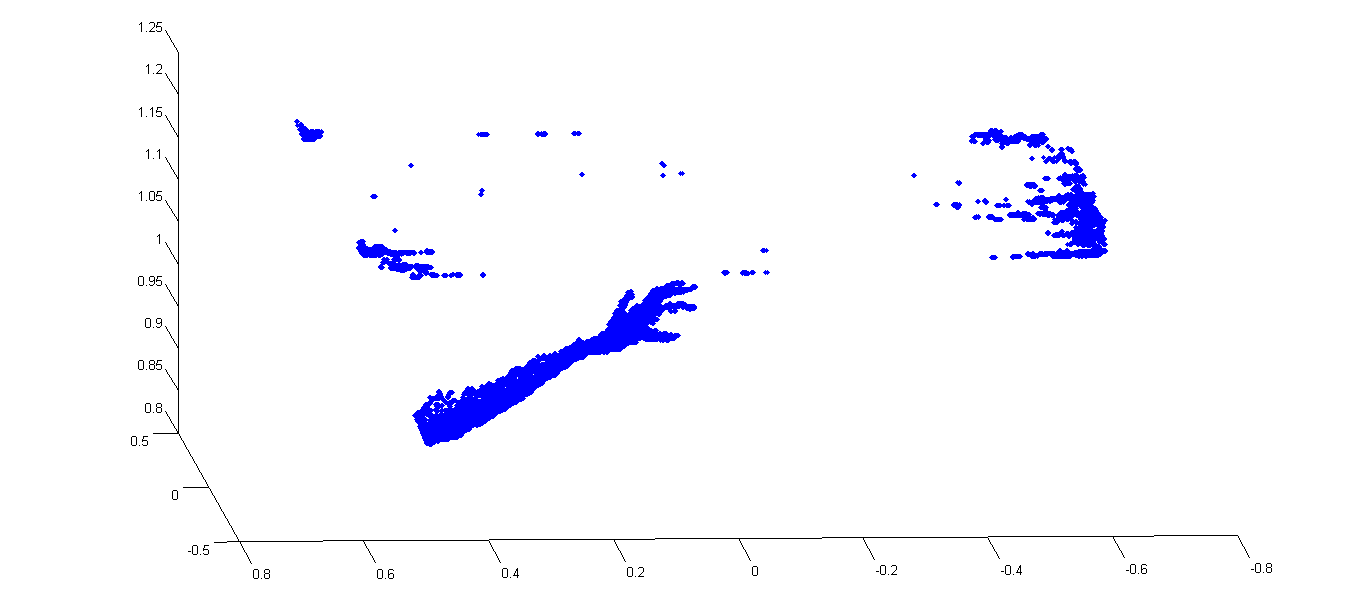}\label{ransac_th4}}
\caption[]{Results of applying RANSAC plane fitting algorithm with different threshold values: (a) Original Point Cloud; b) $\tau = 0.16$; c) $\tau = 0.10$; d) $\tau = 0.01$.}
\label{ransac_th1234}
\end{figure}

The algorithm is described in detail in \citep{Fischler}. For all experiments, we set the maximum number of iterations and the threshold $\tau$ to 1000 and 0.1 respectively. Fig. \ref{ransac_result123} shows the result of the RANSAC algorithm during a grasping task. The background is completely removed without affecting the remaining point cloud represented by the hand and the object. Fig. \ref{ransac_th1234} shows the results of the algorithm using different values of $\tau$. A high threshold means that there is high sensitivity towards the outliers, therefore some points that belong to the hand  are eliminated. A low threshold implies that points that deviate slightly from the plane model are considered as outliers. A proper trade-off considering the distance of the objects from the background must be adopted. 

Background subtraction eliminates much of the data points in the cloud, hence it can be managed more easily and quickly for the next steps where the number of 3-D points typically drops by an order of magnitude. 

\subsection{Segmentation}
\label{subsec:segmentation}
In order to associate each 3-D point to the corresponding cluster in the scene we have tested three different clustering algorithms: K-means, Euclidean Cluster Extraction and DBSCAN. K-means is a popular clustering algorithm that partitions data in a fixed a priori number of clusters. Two steps are iterated recursively until convergence is reached: find \emph{k} centroids as barycenters of the clusters and associate each point to the nearest centroid. \\The Euclidean Cluster Extraction \citep{RusuDoctoralDissertation} is a simple algorithm for clustering a 3-D point cloud based on the Euclidean distance proposed by Rusu. For each point of the data set the algorithm, basically, searches for the nearest points inside a sphere of a fixed radius $r$. The number of clusters is found by processing all the points of the data set whereby only the radius of the sphere is required as an input parameter. 

\begin{figure}
\centering
\subfigure[]{\includegraphics[width=75mm,height=35mm]{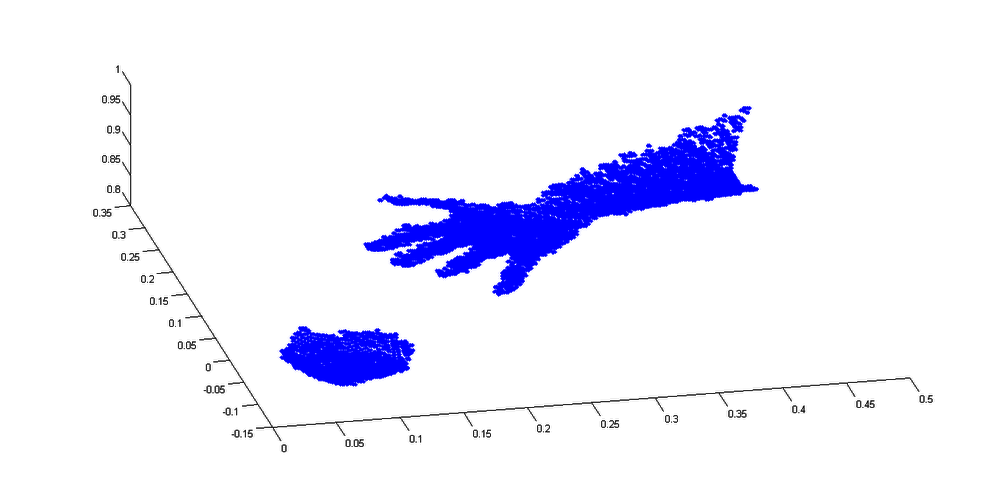}\label{fig:clustering}}\hspace{-1.4em}
\subfigure[]{\includegraphics[width=75mm,height=35mm]{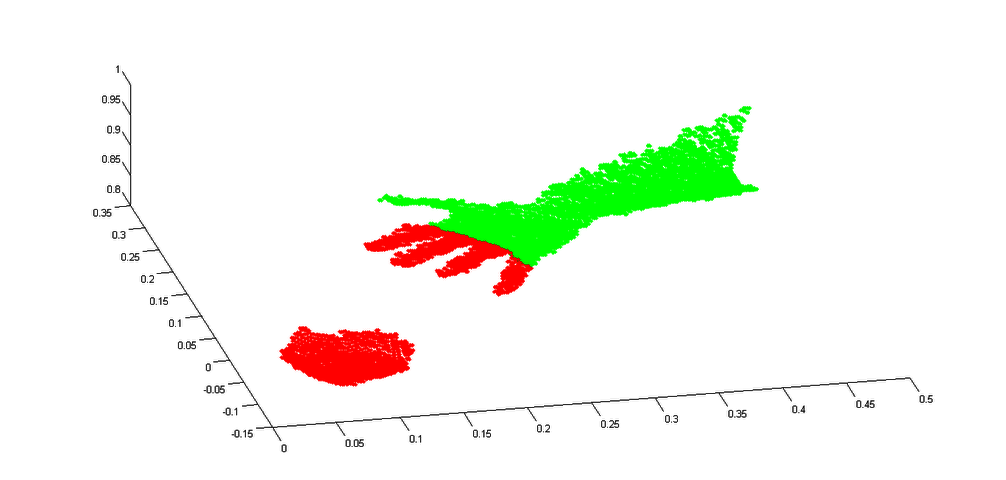}\label{fig:kmeans}}
\subfigure[]{\includegraphics[width=75mm,height=35mm]{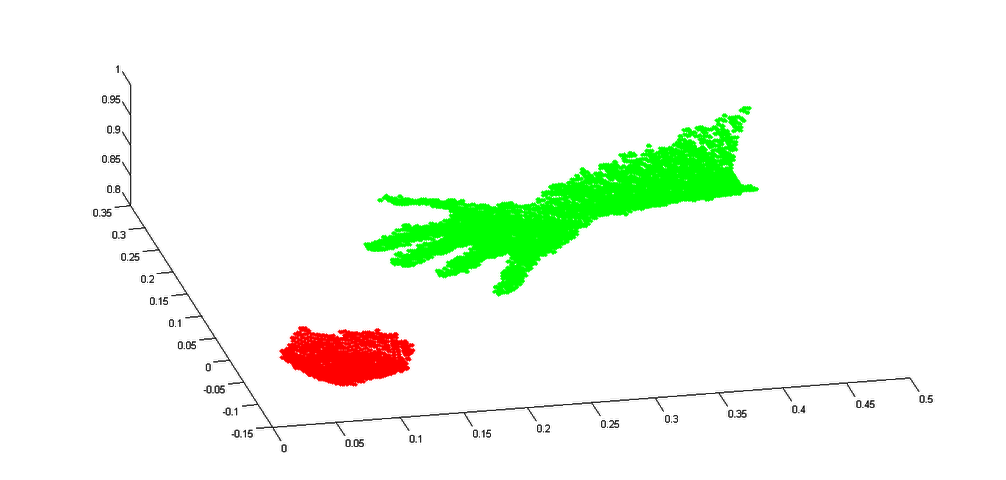}\label{fig:ec}}\hspace{-1.4em}
\subfigure[]{\includegraphics[width=75mm,height=35mm]{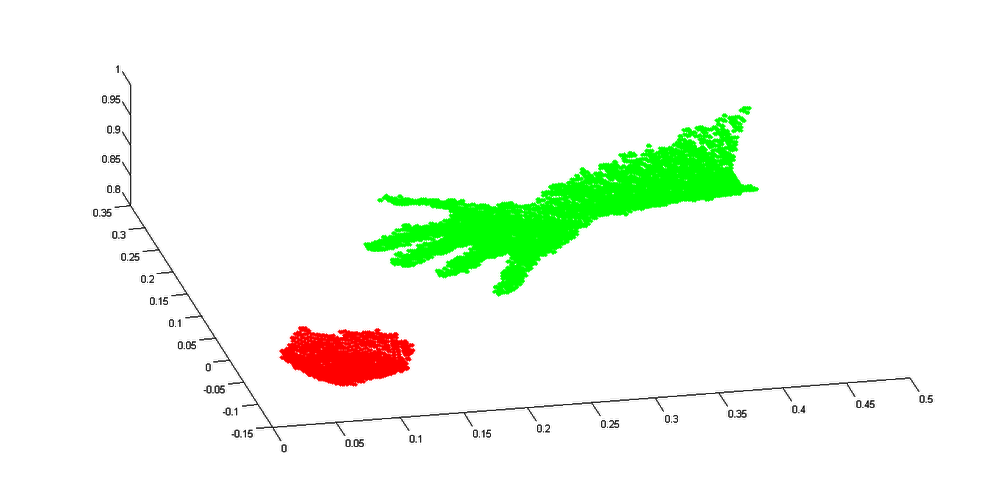}\label{fig:dbscan}}
\caption[Segmentation]{a) Original Point Cloud; Output of the segmentation step by: b) K-means; c) Euclidean Cluster Extraction; d) DBSCAN.}\label{segmentation}
\end{figure}

DBSCAN (Density Based Spatial Clustering of Application with Noise) \citep{Ester96adensity-based} is another technique based on density estimation for arbitrary shaped clusters in spatial databases. The algorithm is based on two notions: cluster and noise. DBSCAN starts with an arbitrary starting point in the dataset that has not been visited. If its \emph{eps}-neighborhood contains a sufficient number of points, it is classified as a core point, otherwise it is labeled as noise. The search continues for all the points in the dataset. DBSCAN is able to differentiate border points from noisy points of a cluster. In fact, if the minimum number of points is recognized later, points previously marked as noise are renamed. The algorithm requires two parameters as input: \emph{eps}, the radius of the sphere, and \emph{minPts}, the minimum number of points to mark a cluster as a core object.\\In both the latter two algorithms, the nearest neighbor searches have been performed creating a kd-tree for each point cloud. We have used the structure written by Tagliasacchi \citep{Tagliasacchi} obtaining a substantial performance improvement for search operations in the point clouds.
Fig. \ref{segmentation} shows the results of the segmentation step on a point cloud during a grasping task. More generally, if two clusters are close enough, K-means it is not able to divide the objects correctly, whereas the DBSCAN and the Euclidean Cluster Extraction algorithms perform much better. In the following steps we have used the results of the DBSCAN algorithm  that in our experiments has appeared to show the best performance.

\subsection{Synthetic hand models generation}
\label{sec:hand_models}
\begin{figure}
\centering
\subfigure[]{\includegraphics[scale=0.16]{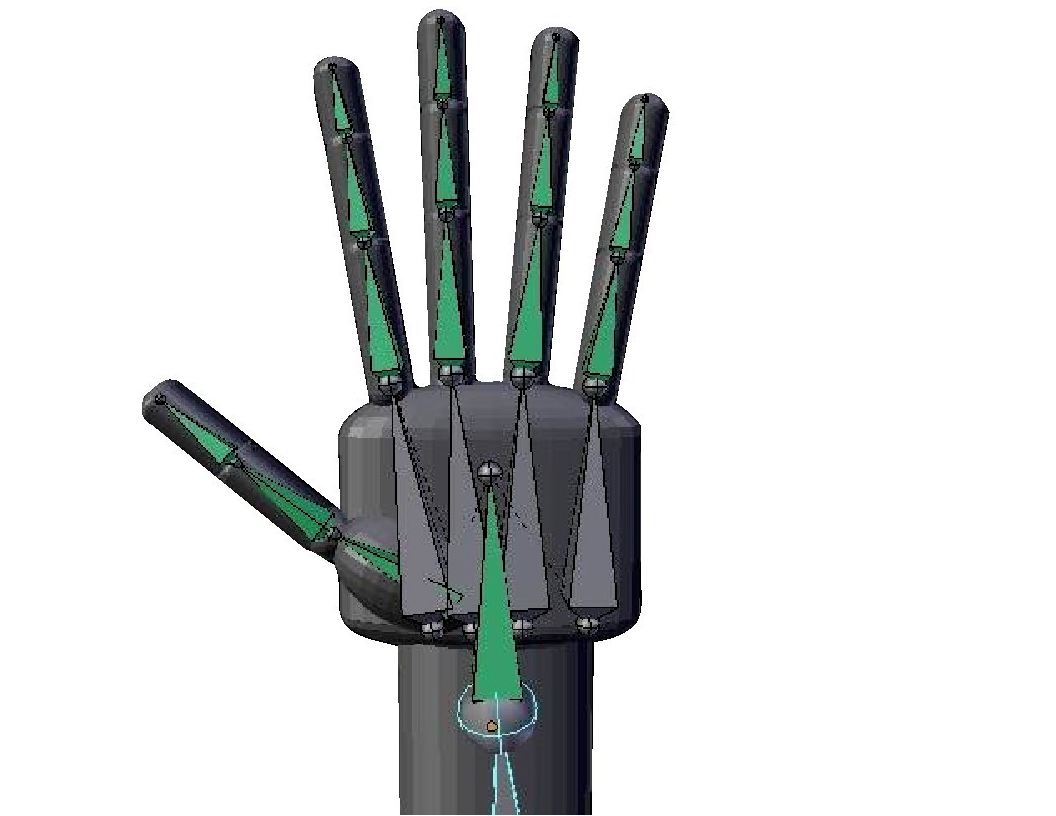}}\hspace{1.5em}
\subfigure[]{\includegraphics[scale=0.23]{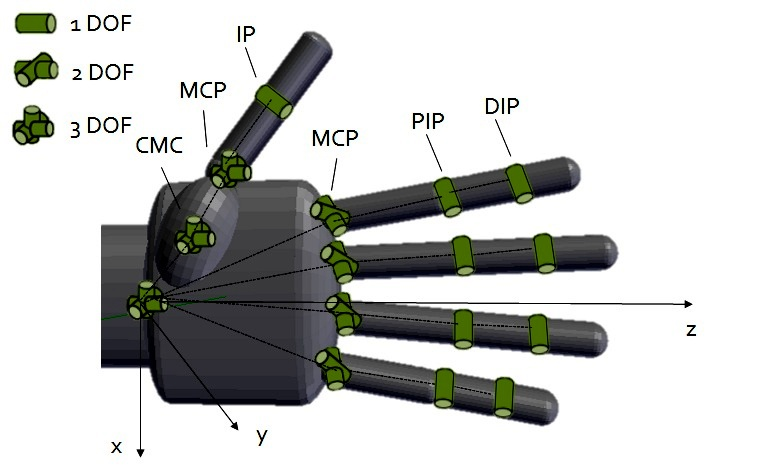}}
\caption[3dhand_model]{a) 3-D hand model and corresponding skeleton that controls its deformations  (constrained bones are shown in green); b) DOFs of the hand joints.}
\label{3dhand_model}
\end{figure}

\begin{figure}
\centering
\subfigure{\includegraphics[scale=0.15]{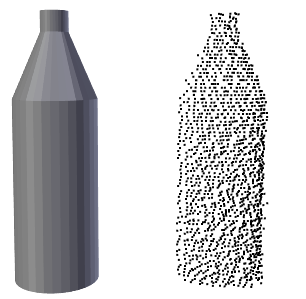}}
\subfigure{\includegraphics[scale=0.15]{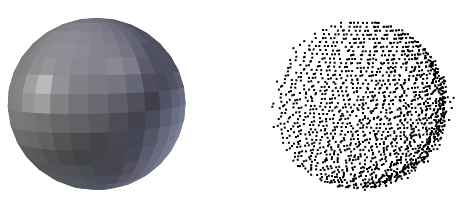}}
\subfigure{\includegraphics[scale=0.11]{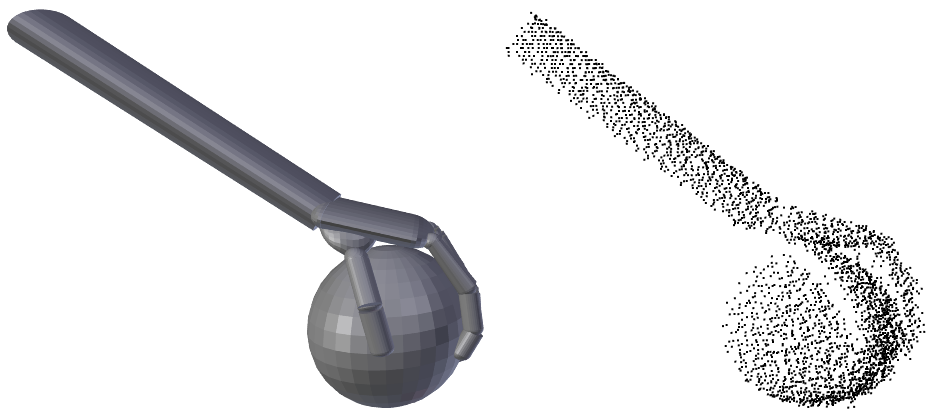}}
\subfigure{\includegraphics[scale=0.13]{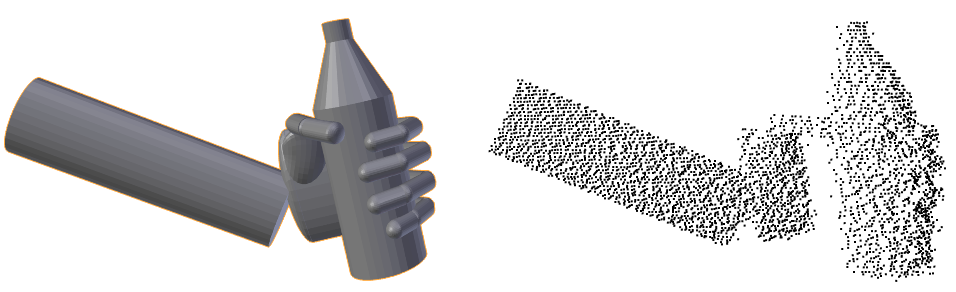}}\\
\subfigure{\includegraphics[scale=0.13]{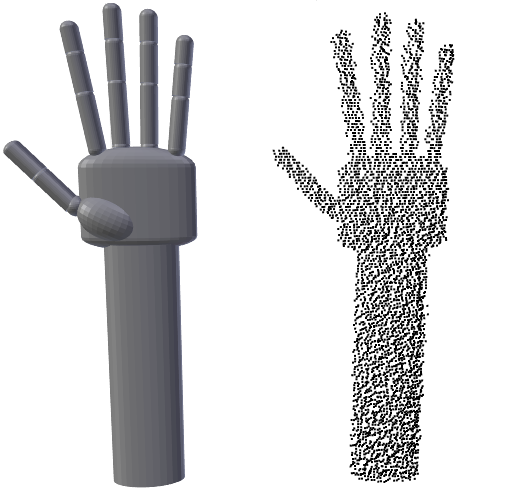}}
\subfigure{\includegraphics[scale=0.13]{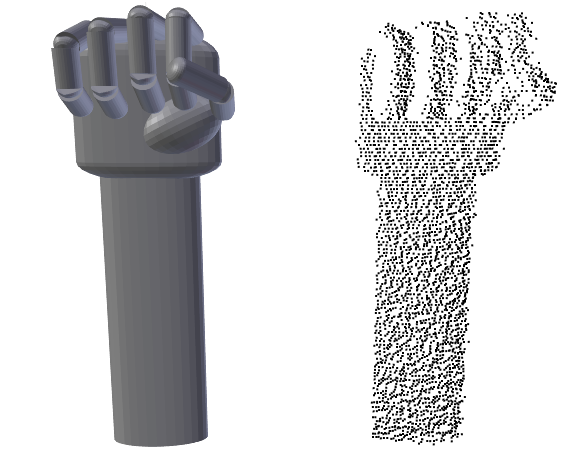}}
\subfigure{\includegraphics[scale=0.13]{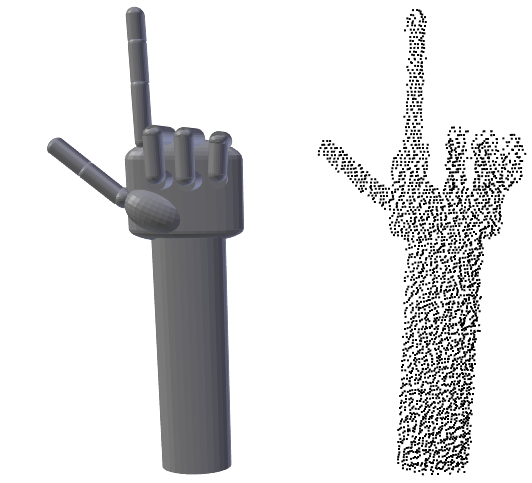}}
\caption[database]{The 3-D models and their corresponding point clouds generated for our dataset.}
\label{database}
\end{figure}

The point clouds prototypes  are obtained from the BlenSor software \citep{Gschwandtner:2011}, which allows the creation of a point cloud from a 3-D model. Blensor is a simulation software based on Blender that simulates the output of various types of sensors such as LIDAR, Time-of-flight (ToF) and Kinect cameras. A 3-D hand model with 26 DoFs has been built by assembling geometric primitives such as cylinders, spheres and ellipsoids. Each of the four fingers has 4 DoFs while the thumb has 7 DoFs as a result of a different structure. The fingers together have 23 DoFs. The remaining 3 DoFs are from the rotational motion of the palm. The shape parameters of each object are set by taking measurements from a real hand. In order to manipulate such model, a skeleton based on \emph{armature} objects is created with rigging process. Rigging is the process of attaching a skeleton to a mesh object that allows to deform and to pose it in different ways. Moreover, to remain within feasible movements, hand and finger motion uses \emph{static} constraints that typically limit the motion of each finger \citep{897381}, see Fig. \ref{3dhand_model}. The constraints, used to obtain the point cloud of the relative pose, are the following:
\begin{equation}
\begin{split}
0^\circ \leq \theta_{MCP-F}\leq 90^\circ,\\
0^\circ \leq \theta_{PIP}\leq 110^\circ,\\
0^\circ \leq \theta_{DIP}\leq 90^\circ,\\
-15^\circ \leq \theta_{MCP-AA}\leq 15^\circ.
\end{split}
\end{equation}
In Fig. \ref{database} some of the 3-D models are shown with their associated point clouds that have been used in building the prototypes for our dataset. In particular, we have considered three typical gestures of the hand (palm, fist and gun) and two ``interaction" models when the hand interacts with an object. 

\subsection{Matching}
\label{subsec:matching}
\begin{figure}
\centering
{\includegraphics[width=75mm,height=35mm]{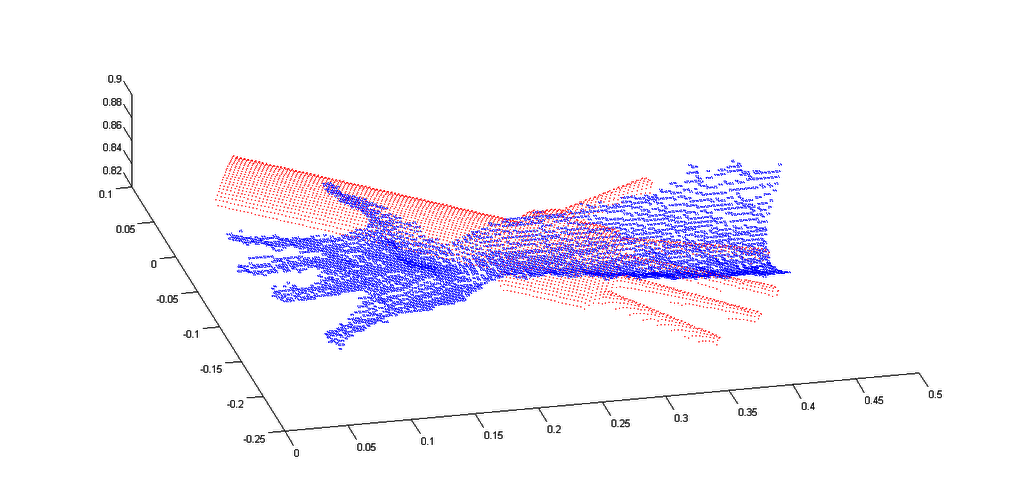}\label{fig:stoch_icp1}}\hspace{-1.4em}
{\includegraphics[width=75mm,height=35mm]{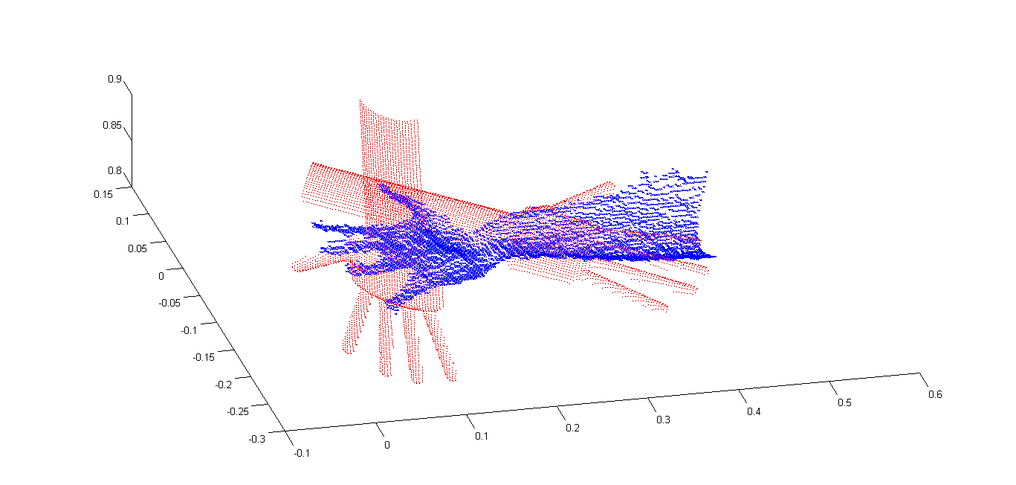}\label{fig:stoch_icp2}}
{\includegraphics[width=75mm,height=35mm]{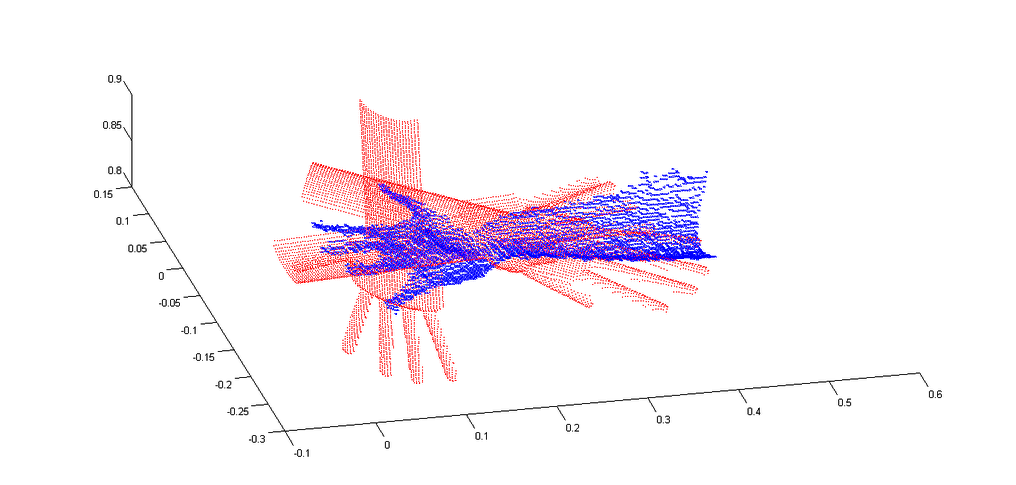}\label{fig:stoch_icp3}}\hspace{-1.4em}
{\includegraphics[width=75mm,height=35mm]{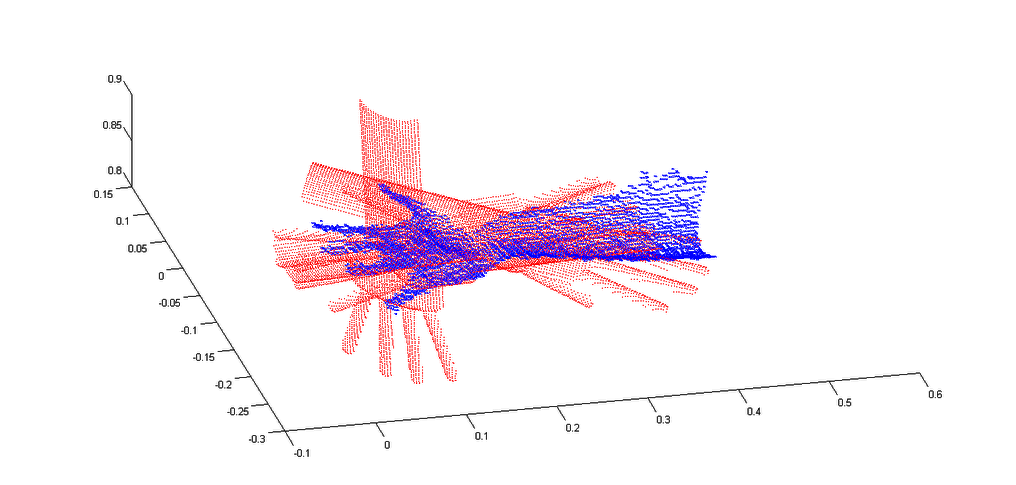}\label{fig:stoch_icp4}}
\caption[Stochastic ICP algorithm]{First four perturbed models obtained by applying the ``stochastic" version of the ICP algorithm. Among all perturbed models, the closest one to the data captured by the sensor is chosen.}
\label{stoch_alg}
\end{figure}

The next step consists of finding the optimal registration between the point clouds in the dataset and the point cloud obtained by the clustering algorithm. To perform the matching between the models and the data acquired by the Kinect sensor, the ICP algorithm \citep{Besl:1992} has been used. In this regard, we provide a detailed description of our system in two typical cases: the hand performs simple gestures and the hand interacts with an object. Due to many problems, such as local minima, noise and partial overlap, the ICP algorithm may converge slowly or not converge at all. In addition, the position of the hand in the space is not known at the beginning of the simulation. For these reasons, we propose a modification of the ICP algorithm considering its ``stochastic" version. \\The goal of the ICP algorithm is to bring two point clouds as close as possible to each other. The algorithm at each iteration step, selects the nearest points of a cloud (the model, in our case) respect to the other one (the input data) and calculates the transformation, represented by a rotation matrix $\textbf{R}$ and a translation vector $\textbf{t}$, that minimize the following equation:
\begin{equation}
E(\textbf{R},\textbf{t})=\sum_{i=1}^{N_p}\sum_{j=1}^{N_m}\textit{w}_{i,j}\|\textbf{p}_i-(\textbf{R}\textbf{m}_j+\textbf{t})\|^2,
\end{equation}
where $\textbf{M}=\{m_1, \ldots, m_{N_m}\}$ is the model's point cloud, $\textbf{P}=\{p_1, \ldots, p_{N_p}\}$ is the input point cloud after the segmentation step and $\textit{w}_{i,j}\in\{0,1\}$ represents the point correspondences. This equation represents the objective function that is minimized at each iteration; in particular, it represents the sums of squared distance of the model points to the data points (known as \emph{point-to-point minimization}). A closed form solution for this rigid body transformation can be calculated using a method based on the singular value decomposition (SVD). The centroids and the deviations from such centroids are given by:
\begin{gather}
\mu_m=\frac{1}{N_m}\sum_{i=1}^{N_m}\textbf{m}_i, ~~~~~~  \mu_p=\frac{1}{N_p}\sum_{i=1}^{N_p}\textbf{p}_i,\\
\textbf{M'}=\{\textbf{m}_i-\mu_m\}=\{\textbf{m}_i'\}, ~~~~~~\textbf{P'}=\{\textbf{p}_i-\mu_p\}=\{\textbf{p}_i'\}.
\end{gather}
Considering the SVD of the matrix \textbf{N},
\begin{equation}
\textbf{N}=\sum_{i=1}^{N_m}{\textbf{p}_i' \textbf{m}_i'^{T}}=U\Sigma V^{T},
\end{equation}
where $U, V \in \mathbb{R}^{3x3}$ are unitary and $\Sigma$ is the matrix of the singular values of \textbf{N}, if $rank(\textbf{N})=3$, the optimal solution of E(\textbf{R}, \textbf{t}) is unique and given by:
\begin{gather}
\textbf{R}=\textbf{U}\textbf{V}^{T},\\
\textbf{t}=\mu_p-\textbf{R}\mu_m .
\end{gather}
The minimal value of the error function is
\begin{equation}
E(\textbf{R}, \textbf{t})=\sum_{i=1}^{N_m}{(\|\textbf{p}_i'\|^2+\|\textbf{m}_i'\|^2)}-2(\sigma_1+\sigma_2+\sigma_3),
\end{equation}
where $\sigma_1\geq\sigma_2\geq\sigma_3$ are the singular values of \textbf{N}. Several methods have been proposed to increase the robustness of the ICP algorithm \citep{icp1, icp2, icp3}. Our approach consists in adding a normally distributed noise and a random rotation matrix to perturb the position of the model's point clouds in order to obtain a good alignment at the first frame (a similar approach was proposed in \citep{stoch_icp}). More specifically: $\textbf{m}_j'=\textbf{R}_{rand}\textbf{m}_j + \textbf{s},\hspace{0.3cm} j = 1, \ldots, N_m$, where $\textbf{s}$ has zero mean and diagonal covariance matrix $\sigma^2 I$ and $\textbf{R}_{rand}$ is a random rotation matrix. The parameter $\sigma$ has been set to 0.1 for the experiments and its value has been determined experimentally. We have used the algorithm proposed by Arvo \citep{arvo} to generate a random rotation matrix. The stochastic version of the ICP algorithm operates as reported in Algorithm \ref{alg:stochastic_icp}: the model \textbf{M} is perturbed as discussed before, and then it is applied the standard version of the ICP algorithm on such perturbed model with respect to the input data \textbf{P}. This operation is repeated a fixed number of times. Among all perturbed models, the one closest to the data captured by the sensor is chosen. Finally, the standard ICP algorithm is repeated on such model using as stopping criteria the two parameters \emph{max\_iter} and \emph{err}. The input parameters $n$ and $err$ have been set respectively to 50 and 1e-5. In Fig. \ref{stoch_alg} the first four perturbed models obtained by applying the stochastic ICP algorithm  are shown.
It is important to notice that this procedure can be parallelized because the model is perturbed using different values of $\textbf{s}$ and $\textbf{R}_{rand}$ every time. 

\begin{algorithm}
\caption{Stochastic ICP}
\label{alg:stochastic_icp}
\begin{algorithmic}[1]
\State \textbf{Input}: {M, P, n, err} \LComment{P is the input point cloud}
    \For{i $\gets$ 1 \textbf{to} n} 
           \State max\_it = 10; 
	\State generate noise s;
	\State generate random matrix \textbf{$R_{rand}$};
	\State $M_i$= $R_{rand}$ $\cdot$ M + s;
	\State [$M_i$, $R_1$, $t_1$]=standard\_ICP($M_i$, P, max\_it, err);
\EndFor
	\State max\_iter = 30;
	\State M = $\arg\min\limits_{M_i}$ dist($M_i$, P);
	\State [$M_{best}$, $R_2$, $t_2$]=standard\_ICP(M, P, max\_iter, err);	
\State \textbf{Output}: $M_{best}$, $R_2, t_2$
\end{algorithmic}
\end{algorithm}

\subsubsection{Matching strategy}
\label{subsec:matching_strategy}
Using the stochastic ICP algorithm presented above, two different strategies have been developed. When the hand does not interact with an object we adopt the following strategy (Algorithm \ref{alg:matching_strategy_1}): in the first frame, the stochastic  ICP algorithm is performed, using the three static gestures (palm, gun, fist) in the dataset. Then, from the model that presents the best matching with the acquired data, the translation vector and the rotation matrix are obtained. All models are updated with these two parameters because they have been created using the same viewpoint and the same position. Starting from the next frame, to the end of simulation, the standard ICP algorithm is performed since it converges for relatively small translations and rotations around all axes. In fact, we can reasonably assume that the hand executes small movements between two consecutive frames. All models are still updated with the pose estimated in the previous frame.
\\When the hand interacts with an object, we assume to begin with two separate clusters. In this case, we need an initial phase of detection. Firstly, the stochastic ICP algorithm is applied to determine which cluster is the object by performing the matching algorithm between the objects' point clouds in the dataset and each of the two clusters. The cluster with the minimum error is labeled as object ($C_1$); therefore, the other cluster is labeled as hand  ($C_2$). Then, the stochastic ICP algorithm is repeated on the hand cluster ($C_2$) using the three static hand gestures in the dataset to determine the hand pose. As long as the two clusters are separated, the standard version of the ICP algorithm is applied, using the three models for the hand cluster and the detected object's model for the other one to update the pose of both models. When the clusters are merged into a single one, the stochastic ICP algorithm is applied using only the two ``interaction'' models in the dataset. In this case, the stochastic ICP allows to obtain a more robust matching. The strategy is reported in detail in Algorithm \ref{alg:matching_strategy_2}.

\begin{algorithm}
\caption{Matching strategy for static gestures}
\label{alg:matching_strategy_1}
\begin{algorithmic}[1]
\State \textbf{Input} : {$M_i$, P} \LComment{P is the input point cloud}
\If{first frame}
	\For{all static gestures models}
		\State [$M_i$, $R_i$, $t_i$] = stochastic\_icp($M_i$, P)
	\EndFor
\Else { \LComment{This is not the first frame}}
	\For{all static gesture models}
		\State $M_i =  R_{best} \cdot M_i + t_{best}$; \LComment{pose update}
		\State [$M_i$, $R_i$, $t_i$] = standard\_icp($M_i$, $P$);
	\EndFor 
\EndIf
\State $[M_{best}, R_{best}, t_{best}]$ = $\arg\min\limits_{M_i}$ dist($M_i$, P); \LComment{Find the best matching model}
\end{algorithmic}
\end{algorithm}
\begin{algorithm}
\caption{Matching strategy for grasping gestures}
\label{alg:matching_strategy_2}
\begin{algorithmic}[1]
\If{first frame} \LComment{Hand and object detection phase}
 	\\\LComment{Detect the object} 
 	\State [$M_i$, $R_i$, $t_i$] = stochastic\_icp($M_i$, $C_{1,2}$); \LComment{For all objects models}
	\\\LComment{Detect the hand pose}
	\State [$M_i$, $R_i$, $t_i$] = stochastic\_icp($M_i$, $C_2$); \LComment{For all static gestures}
\Else {\LComment{This is not the first frame}}
	\If{clusters are not merged}
	\State [$M_i$, $R_i$, $t_i$] =  standard\_icp($M_i$, $C_{1,2}$); \LComment{For both clusters separately}
	\Else{\LComment{Clusters are merged as $P$}}
	\State [$M_i$, $R_i$, $t_i$] = stochastic\_ICP($M_i$,$P$); \LComment{Using  the ``interaction'' models }
	\EndIf
\EndIf
\State Obtain $R_{best}$ and $t_{best}$ of the best matching models for the two clusters
\State $M_i = R_{best} \cdot M_i  + t_{best}$; \LComment{Pose update for all models}
\end{algorithmic}
\end{algorithm}

\section{Experimental results}
\label{sec:exp_results}
\begin{figure}[]
  \centering
  \includegraphics[scale=0.4]{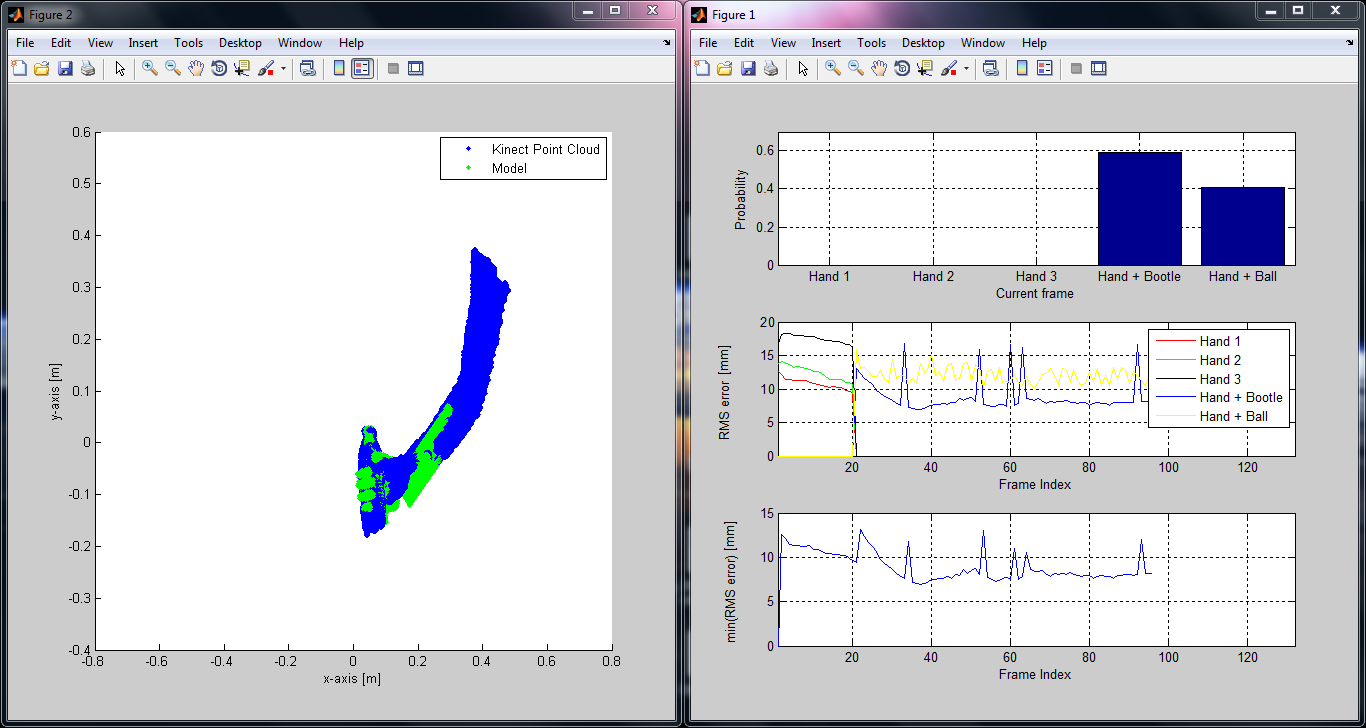}\\
  \caption{Interface of the proposed hand pose estimation system. Left side: result of the matching step. Right side: probability and RMS error of each model and RMS error of the selected model. The model with the highest probability is plotted (in green) in the same figure of the clustered input data.}\label{fig:interface}
  
\end{figure}

\begin{figure}[!t]
\centering
\subfigure[]
{\includegraphics[scale=0.46]{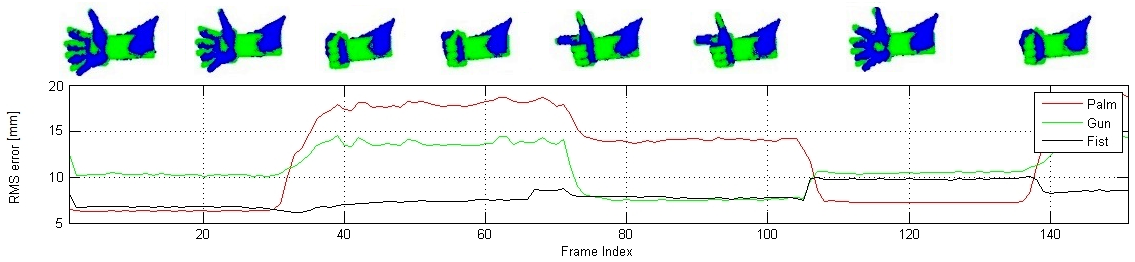}\label{fig:exp_1}}
\subfigure[]{\includegraphics[scale=0.46]{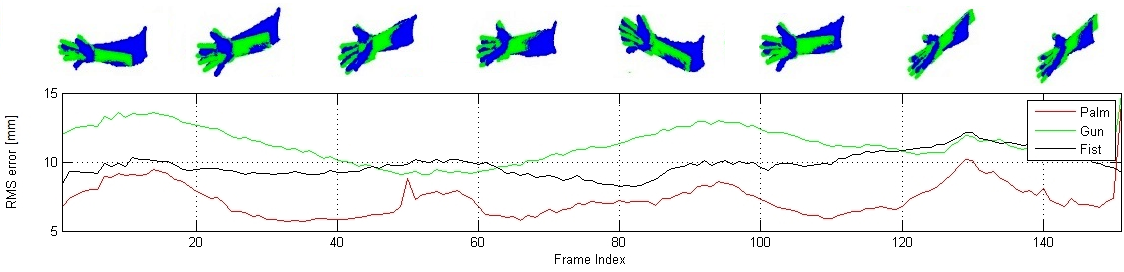}\label{fig:exp_2}}
\subfigure[]{\includegraphics[scale=0.46]{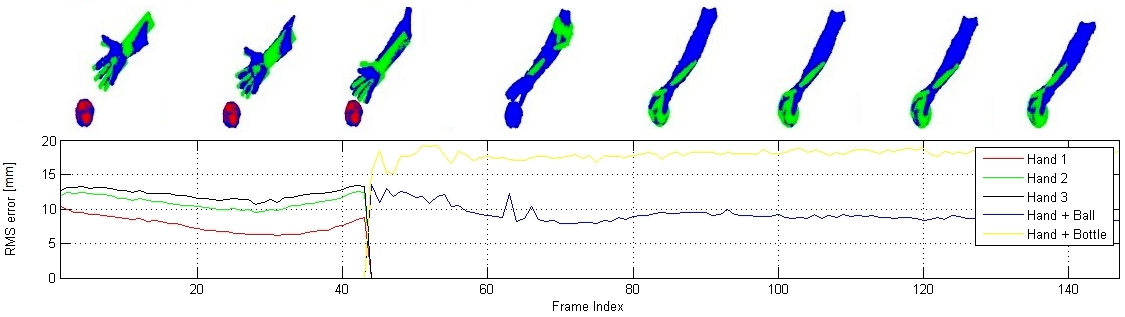}\label{fig:exp_3}}
\subfigure[]{\includegraphics[scale=0.46]{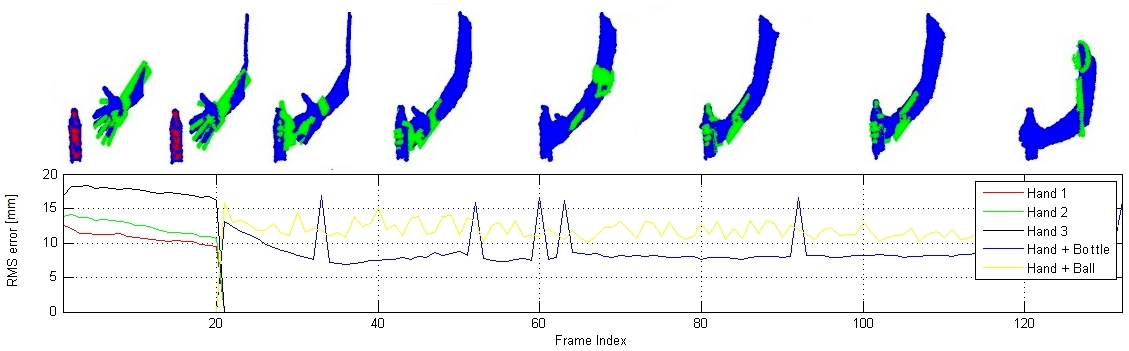}\label{fig:exp_4}}
\caption[]{Sample image sequences of the matching step and the corresponding RMS error: a) Hand performs static gestures; b) Hand performs rotations showing the palm; c) Ball grasping task; b) Bottle grasping task. }
\label{im:qual_results}
\end{figure}

In order to evaluate the accuracy of the system, we have used the RMS error obtained by considering the Euclidean distance between each point of the selected model and the input data after the clustering step. Since the input data typically contain a number of points greater than the models, because of the presence of the arm or the other body parts, this distance is evaluated considering the nearest points of both point clouds. In addition, a measure of probability $p_i[k]$,  of the i-\emph{th} model at the k-\emph{th} frame, has been defined using the corresponding RMS error as follows:  

\begin{equation}
p_i[k] = \frac{\frac{1}{RMSE_i[k]}}{\sum_{i=1}^{N_{models}}\frac{1}{RMSE_i[k]}}.
\end{equation}
\\
Fig. \ref{fig:interface} show the interface of the proposed system. The probabilities are plotted using a bar graph updated every frame.\\ Even though the system does not require an explicit initialization step, we have used an easily recognizable hand pose represented by a palm completely visible at the beginning of simulation. In fact, the system does not perform a real tracking and we considered only a limited number of poses.  In the next two sections, the experimental results of our approach are presented. Fig. \ref{im:qual_results} shows both the qualitative results and the corresponding RMS errors.  

\subsection{Static gestures}
The hand is positioned approximately at a distance of 1.8 \emph{m} from the Kinect sensor. In this case, the system is able to recognize the three models in the dataset. In fact, the RMS error is less than 10 \emph{mm} and it remains constant during all the simulation since the gestures are distinguishable and correspond exactly to the synthetic models. When the hand moves by rotating, the error fluctuates more even if the correct pose is still recognized. In the last part of simulation, the error is greater and  the wrong model  is selected mainly because of a high rotation of the hand that no longer matches the models in the dataset. The visible part, which is captured by the Kinect, is effectively only the profile of the hand and a greater rotation of the model, even if it is correct, results in an increase of the RMS error. 

\subsection{Grasping objects}
Finally, we have tested our system when the hand interacts with an object. In particular, we have considered the grasping of objects easily obtainable in 3-D such as a ball and a bottle. For this experiments, we have replaced the three gestures (palm, gun, fist) with three intermediate poses representing the hand that comes close to the objects. The approach works quite well in the initial phase, recognizing the correct pose of the hand and the objects. As soon as the clustering algorithm returns a single cluster in the scene, only the interaction models are considered. As expected, the approach shows obvious problems in recognizing the correct model, when the clusters begin to join, because they are not completely merged. For the grasping ball task, the RMS error remains less than 10 \emph{mm}, as in the above experiments. However, for some frames, the correct model is chosen but the estimated pose is wrong, primarily because of a lack of a real tracking. For the grasping bottle task some RMS error peaks are visible due to the presence of the entire arm that increases the surface area that needs to be considered for matching. 
\section{Conclusions and future work}
In this work, we have presented a hand pose estimation system which is able to recognize a limited number of hand poses and interaction scenarios combining algorithms widely used in computer vision. The experimental results are very promising and demonstrate the effectiveness of the proposed approach. The system is able to provide a reasonable 3-D hand pose estimation using predefined models, as confirmed by the RMS error, also considering more complex scenarios in which the hand interacts with different objects.\\In our future work, possible improvements will be investigated. First of all, the system performance could be improved using parallelized code on GPU or by means of the use of the Point Cloud Library (PCL) \citep{3dpcl}, an open-source library of algorithms specifically created for point-cloud processing tasks and 3-D geometry processing. Moreover, a larger dataset of typical gestures will be designed. Furthermore, an increase in accuracy could be achieved by implementing a tracking module to support the pose estimation.


\label{conclusion}
\bibliography{mybib}

\end{document}